\documentclass{article}

\usepackage{microtype}
\usepackage{graphicx}
\usepackage{subfigure}
\usepackage{booktabs} 
\usepackage{amsmath}

\usepackage{hyperref}

\usepackage[accepted]{icml2019}

\usepackage{float}

\usepackage{pgfplotstable}

\usepackage{dsfont}

\usepackage{amsfonts}
\usepackage{amssymb}
\usepackage{amsthm}
\usepackage{bbm}
\usepackage{breqn}

\theoremstyle{remark}

\theoremstyle{definition}

\newcommand{\R}{\mathbb{R}}
\newcommand{\E}{\mathbb{E}}
\newcommand{\J}{\mathcal{J}}

\newcommand{\normsq}[1]{\left\lVert#1\right\rVert^2}

\icmltitlerunning{Image Deconvolution via Noise-Tolerant Self-Supervised Inversion}

\begin{document}

\twocolumn[
\icmltitle{Image Deconvolution via Noise-Tolerant Self-Supervised Inversion}



\icmlsetsymbol{equal}{*}

\begin{icmlauthorlist}
\icmlauthor{Hirofumi Kobayashi }{czb}
\icmlauthor{Ahmet Can Solak }{czb}
\icmlauthor{Joshua Batson }{czb}
\icmlauthor{Loic A. Royer \textsuperscript{*}}{czb}
\end{icmlauthorlist}

\icmlaffiliation{czb}{Chan-Zuckerberg Biohub}

\icmlcorrespondingauthor{Loic Royer}{loic.royer@czbiohub.org}

\icmlkeywords{Machine Learning, Inverse Problems, Deconvolution}

\vskip 0.3in
]



\printAffiliationsAndNotice{}  

\begin{abstract}
We propose a general framework for solving inverse problems in the presence of noise that requires no signal prior, no noise estimate, and no clean training data. We only require that the forward model be available and that the noise be statistically independent across measurement dimensions. We build upon the theory of ``$\mathcal{J}$-invariant'' functions \cite{batson2019noise2self} and show how self-supervised denoising \emph{à la} Noise2Self is a special case of learning a noise-tolerant pseudo-inverse of the identity.  We demonstrate our approach by showing how a convolutional neural network can be taught in a self-supervised manner to deconvolve images and surpass in image quality classical inversion schemes such as Lucy-Richardson deconvolution.
\end{abstract}

Inverse problems are a central topic in imaging. Rarely are images produced by microscopes, telescopes, or other instruments unscathed. Instead, they often need to be restored or reconstructed from degraded or indirect measurements. Imperfections such as measurement and quantization noise conspire to prevent perfect reconstruction.

\paragraph{Classical approaches to inversion.}
The classical approach to solving inverse problems in the presence of noise typically requires the formulation of a loss function consisting of a data term that quantifies the fidelity of solutions to observations via the forward model, and a prior term that quantifies adherence of solutions to a preconceived notion of what makes a solution acceptable. Typical priors invoke the notion of sparsity in some basis or require smoothness of the solution \cite{McCann2017Convolutional}. An often used prior in image restoration and reconstruction is the Total Variation (TV) prior. Several algorithms have been proposed to efficiently solve the total variation minimisation problem \cite{chambolle2011first}. However, the strength of classical approaches is also their weakness in that the assumptions inherent to the priors are often simplistic and cannot capture the full complexity of real data.

\paragraph{Convolutional neural networks.}
In recent years, deep convolutional neural networks (CNNs) have been shown to outperform previous approaches for various imaging applications. \cite{belthangady2019applications, McCann2017Convolutional}, including denoising \cite{zhang_beyond_2017}, deconvolution \cite{xu2014deep}, aberration correction \cite{krishnan2020optical}, compressive sensing \cite{mousavi2017learning} and super-resolution \cite{dong2014learning}. It has even been shown that CNNs can learn a natural image prior (in the form of a projection operator) that can be used solve all of the above mentioned linear inverse problems \cite{rick2017one}. Yet, all these methods are based on supervised learning and thus require clean training data which is not always available nor obtainable.

\paragraph{Self-supervised learning.}
More recently, self-supervised learning methods have demonstrated their potential for imaging applications. In general, self-supervised learning refers to training a machine learning model without ground truth, solely on the basis of the observed image's statistical structure. This training modality considerably eases the burden of obtaining clean ground truth data. In some applications, self-supervised learning was shown to attain better performance than its supervised counterparts \cite{he2019momentum, misra2019self}.
Self-supervised learning has been successfully applied to imaging, particularly for image denoising where methods have been proposed that only assume pixel-wise statistical independence of noise \cite{lehtinen2018noise2noise, laine2019high, batson2019noise2self, krull2019noise2void, moran2019noisier2noise}.

\paragraph{Self-supervised inversion.}
Self-supervised learning has been explored to solve inverse problems too. For instance, \cite{zhussip2019training} showed that a CNN model can achieve compressed sensing recovery and denoising without the need for ground truth. Recent work by \cite{hendriksen2020noise2inverse} leverages pixel-wise independence of noise to reconstruct images from linear measurements (e.g. X-ray CT) in cases where the inverse operator is known and well-conditioned. Another approach for self-supervised learning is to use adversarial training. A generative adversarial network (GAN) solely trained on corrupted training data can output clean images \cite{pajot2018unsupervised}. More recent work shows that a composite of several GAN models trained on blurred, noisy, and compressed images can generate images free of any such artifacts \cite{kaneko2020blur}. Yet, since these approaches use generative models, they may hallucinate image details, a dangerous property in a scientific context.

In the following we (i) present a generic theory for noise-tolerant self-supervised inversion based on the framework of $\J$-invariance, (ii) apply it to the problem of deconvolving noisy and blurred images, and (iii) evaluate the performance of our approach against four competing approaches on a diverse benchmark dataset of 22 images.


\section{Theory}\label{section:theory}

\paragraph{Problem statement.}
Consider a measurement of a system with forward model $g$ and stochastic noise $n$. We desire to recover the unknown state $x$ from the observation $y = n \circ g (x)$. In the case where there is no noise, i.e., $n$ is the identity function, this reduces to finding a (pseudo)-inverse for $g$. In the case where $g$ is the identity, this reduces to finding a denoising function for the noise distribution $n$. One general strategy is based on optimization, where a prior on $x$ manifests as a regularizer $W$, and one seeks to minimize a total loss $\normsq{g(x) - y} + W(x)$. This requires one to solve an optimization problem for each observation, and also requires an arbitrary choice of the strength and class of the prior $W$. Alternatively, one want to learn a noise-tolerant pseudo-inverse of $g$, but in the absence of training data $(x, y)$ it is not clear how. If one naively optimizes a self-consistency loss $\normsq{g(f(y)) - y}$, then $f$ may learn to invert $g$ while leaving in the effects of the noise $n$, producing a noisy reconstruction. For example, if $g$ represents the blurring induced by a microscope objective (convolution with the point-spread-function), then setting $f$ to be the corresponding sharpening filter (convolution with the the Fourier-domain reciprocal of $g$) will greatly amplify the noise in $y$ while producing a self-consistency loss of $0$.  We propose a modification of this loss which rewards both noise suppression and inversion.

\paragraph{Proposal.}
We extend the $\mathcal{J}$-invariance framework for denoising introduced in \cite{batson2019noise2self}, which applies in cases where the noise is statistically independent across different dimensions of the measurement. Recall that a function $f: \R^m \rightarrow \R^m$ is $\J$-invariant with respect to a partition $\J = \{J_1, \dots, J_r\}$ of $\{1, \dots, n\}$ if for any $J\in \J$, the value of $f(x)_J$ does not depend on the value of $x_J$ \footnote{where $x_J$ denotes $x$ restricted to dimensions in $J$ }. If a stochastic noise function $n$ is independent across the partition, i.e., $n(x)_J$ and $n(x)_{J^c}$ are independent conditional on $x$, and the noise has zero mean, $\E [n(x)] = x$, then \cite{batson2019noise2self} show that the following holds for any $\J$-invariant $f$:
\begin{dmath}\label{eqn:loss_decomposition}
\E \normsq{f(n(x)) - n(x)} = \E \normsq{f(n(x)) - x} + \E \normsq{x - n(x)}.
\end{dmath}
That is, the self-supervised loss (the distance between the noisy data and the denoised data) is equal to the ground-truth loss (the distance between the clean data and the denoised data), up to a constant independent of the denoiser $f$ (the distance between the clean data and the noisy data).

Now let us consider the case where the noise is applied after a known forward model $g$, so: $y = n \circ g (x)$. We are interested in the following generalised self-supervised loss:
\begin{dmath}\label{eqn:generalised_loss_try}
\E \normsq{g(f(y))) - y}
\end{dmath}

To decompose Eq.~\ref{eqn:generalised_loss_try} similarly to Eq.~ \ref{eqn:loss_decomposition} would require $g\circ f$ to be $\J$-invariant. Unfortunately, in the general case, it is difficult to specify properties of $f$ that would guarantee the $\J$-invariance of $g\circ f$. This makes a strategy of explicit $\J$-invariance, where the architecture of $f$ itself is $\J$-invariant as in \cite{laine2019high}, difficult to pursue. However, a simple masking procedure can turn any function into a  $\J$-invariant function, which will let us leverage $\J$-invariance when computing a training loss, even if the final function used for prediction is not itself $\J$-invariant.

Given the partition $\J$, we choose some family of masking functions $m_J$. For example, $m_J$ could replace coordinates in $J$ with zeros, by random values, or by some interpolation of coordinates outside of $J$. Then, for any function $f$ and our fixed forward model $g$, we then compute the following loss:

\begin{dmath}\label{eqn:generalised_loss}
\E \sum_J \normsq{(g\circ f \circ m_J)(y)_J - y_J}.
\end{dmath}

Because the composite function $h$ defined by 
\begin{dmath}
    h_J = (g\circ f \circ m_J)_J
\end{dmath}
is $\J$-invariant, Equation~\ref{eqn:loss_decomposition} applies, and the loss is equal to
\begin{dmath}\label{eqn:generalised_loss_decomposition}
\E \sum_J \normsq{(g\circ f \circ m_J)(y)_J - g(x)_J} + 
\E \normsq{y - g(x)}.
\end{dmath}
As before, the first term is a ground-truth loss (comparison of the noise-free forward model applied to the reconstruction to the noise-free forward model applied to the clean image) and the second term is a constant independent of the pseudo-inverse $f$.

In this formulation, learning a denoising function in \cite{batson2019noise2self} is the special case of learning a noise-tolerant inverse of the identity function.

\paragraph{Differential learning.} Consider  $f_\theta$, a $\theta$-parameterized family of differentiable functions from which we aim to find the best noise-tolerant inverse $f_{\hat{\theta}}$. Since the loss in Eq.~\ref{eqn:generalised_loss_decomposition} is defined in terms of $h_J$, and not in terms of $f$, we need a scheme to optimize $f_\theta$ through the fixed forward model $g$. Assuming that the forward model is also differentiable, we propose to solve this optimization problem by stochastic optimization and gradient backpropagation.

In the following we show how this framework can be used to deconvolve images in a noise-tolerant manner by implementing the forward model with a convolution, and the pseudo-inverse with a Convolutional Neural Network.

\section{Application}
\begin{figure*}[!ht]
  \centering
  \includegraphics[scale=1]{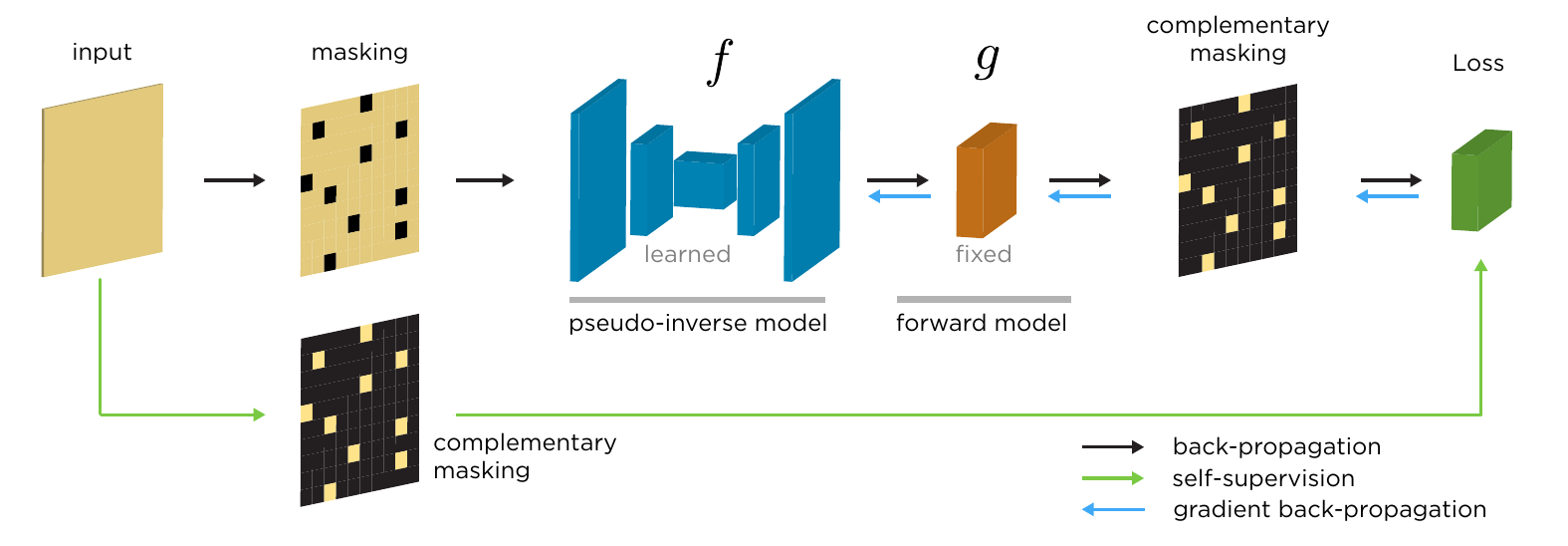}
  \caption{Training strategy for Self-Supervised Inversion. To learn a noise-tolerant pseudo-inverse of a forward model $g$, we train the composite model $g \circ f$ to be a denoiser using a generalised self-supervised loss (Eq.~\ref{eqn:generalised_loss}). We feed observed images $y$ as input and force the network to learn to return back $y$. First, the model transforms a masked observation $y$ into a candidate deconvolved image $f(m_J(y))$. Second, this candidate deconvolved image is passed through the fixed forward model $g$ to return back an observation, which is compared to the original observation on the previously masked pixels, and the error is used to update $f$ by backpropagation. At inference time, we deblur and denoise an image $y$ simply by applying $f$ to it. In the absence of masking, $g \circ f$ might learn the identity function, and the deconvolution $f$ would suffer from noise.}
  \label{fig:strategy}
\end{figure*}
\paragraph{Deconvolving noisy images.} To demonstrate our framework we apply it to the standard inverse problem of image deconvolution. In this case the forward model $g$ is the convolution of the true image $x$ with a blur kernel $k$. The observed image $y$ is thus:
\begin{dmath}\label{eqn:imaging_model}
y = n(k * x)
\end{dmath}
In the case that the noise function $n$ is the identity, the problem can be solved perfectly\footnote{Assuming compact support for $k$ and infinite numerical precision.} by using the inverse filter $k^{-1}$. However, in general and in practice many measurement imperfections such as signal quantization, signal-dependent, and signal-independent noise conspire to make $n$ far from the identity. In the simulations in this paper, we use a Poisson-Gaussian noise model augmented with `salt\&pepper', a good model for low signal-to-noise observations on camera detectors. We also focus on the 2D case, though the architecture and argument work in arbitrary dimension.

\begin{figure}[!ht]
  \centerline{\includegraphics[width=\columnwidth]{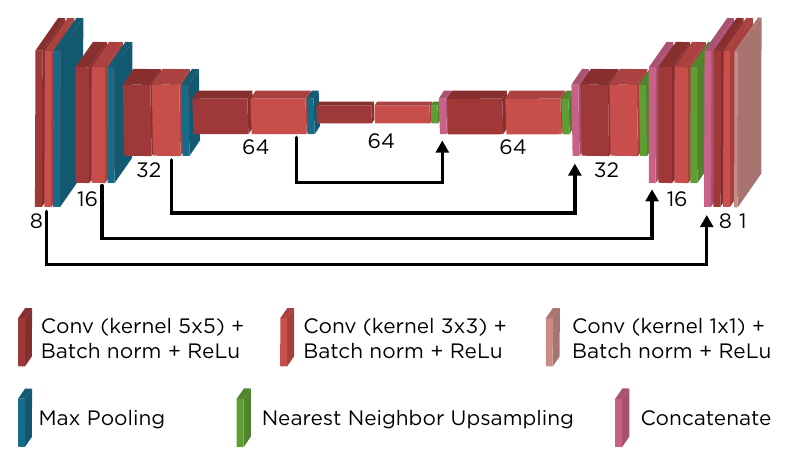}}
  \caption{Detailed model architecture for the UNet used for $f$. The numbers correspond to the number of channels in each layer.}
  \label{fig:architecture}
\end{figure}

\paragraph{Training strategy and model architecture.}
Fig.~\ref{fig:strategy} illustrates our self-supervised training strategy. Instead of a single trainable model $f$ as in Noise2Self \cite{batson2019noise2self} we train the composition of a trainable inverse $f$ followed by the fixed forward model $g$ under the generalised self-supervised loss in Eq.~\ref{eqn:generalised_loss}. As shown in Fig.~\ref{fig:architecture} we implement $f$ with a standard UNet \cite{ronneberger_u-net:_2015}. Once the function $g\circ f$ has been trained, we can use $f$ as pseudo-inverse to deconvolve the blurred and noisy image $y$. The use of a masking scheme guarantees noise-tolerance during training. Yet, since the forward model $g$ is effectively a low-pass filter, it is conceivable that the model $f$ may introduce spurious high frequencies that would then be suppressed by $g$ and thus never be seen nor penalised by the loss. Due to the stochastic nature of neural network training we observe occasional failed runs that lead to noticeable checker-board artifacts. The choice of nearest neighbour up-scaling in the UNet does alleviate this problem significantly. We have also experimented with a kernel continuity regularisation scheme that favours smoothness in the last convolution kernels of the UNet -- but this has a cost in terms of sharpness. In practice, we are pleased by the absence of strong ringing artifacts -- probably because of the combination of convolutional bias \cite{ulyanov2018deep} and our usage of weight regularisation that penalises the generation of unsubstantiated details (both $L_1$ and $L_2$ regularisation, see code for implementation details). Finally, we found that starting with a high masking density of 50\% and slowly decreasing it as epochs progress down to 1\% helps with training efficiency. Starting with a high masking density helps training to get started for low frequencies first. As the masking density decreases, training efficiency and thus speed decreases too, but that also means more
accurate training, because masking disrupts training in itself. 

\section{Results}

\begin{figure*}[!ht]
  \centering
  \includegraphics[scale=1]{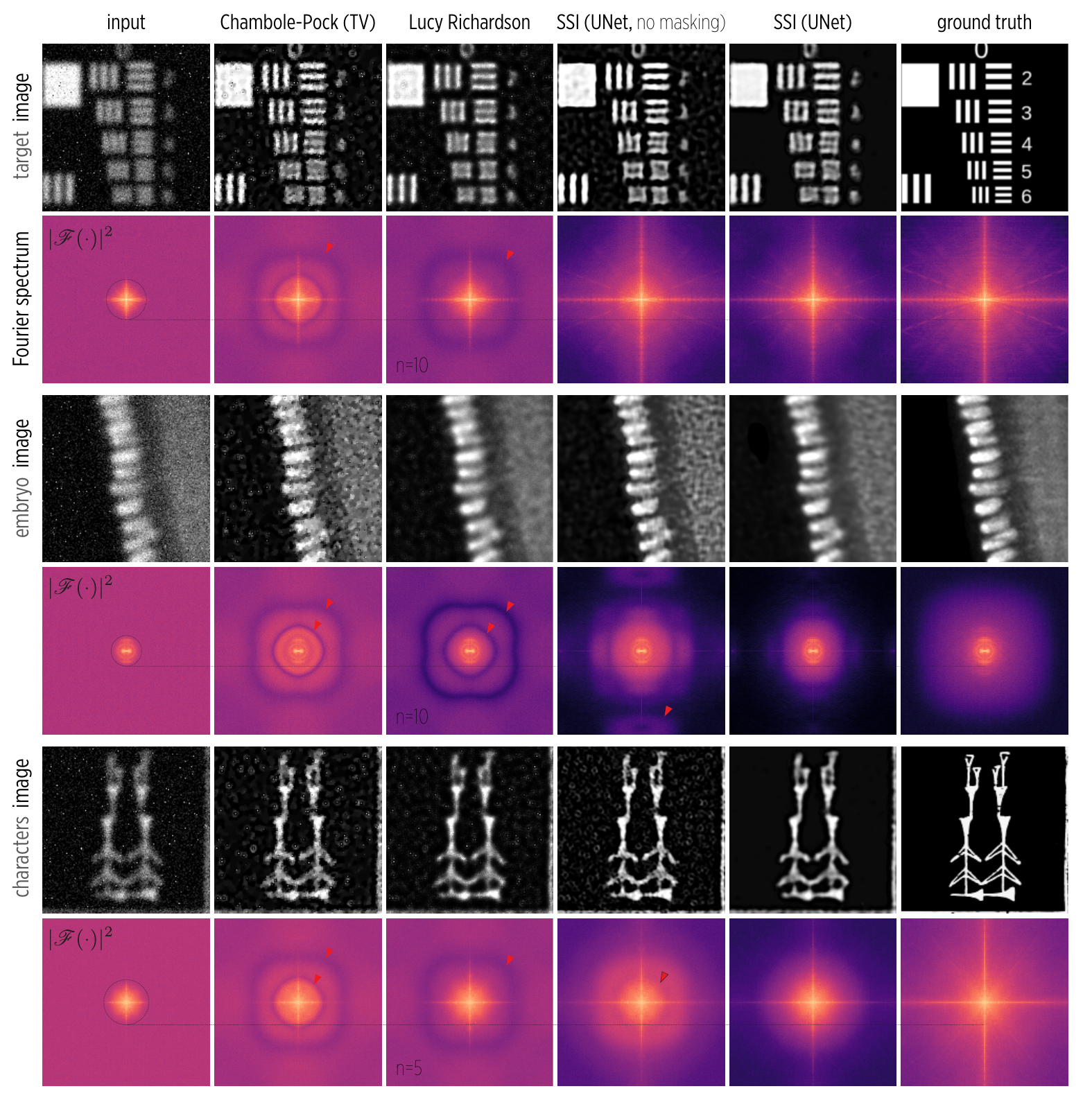}
  \caption{Performance of classic, and self-supervised inversion methods on natural images on three selected images. We show crops (80x80 pixels) as well as whole image spectra for four approaches: Chambole-Pock (CP) with a TV prior, Lucy-Richardson (LR), and Self-Supervised Inversion (SSI) via UNet with, and without, masking. SSI reconstructions achieve a good trade-off between noise reduction and high spatial frequency fidelity. The classic methods (CP, LR) are more likely to introduce distortion at high frequencies (red pointers) whereas SSI spectra a rather clean at high frequencies. Moreover, SSI spectra have a low noise-floor which corresponds to good noise reduction. However, the masking procedure which introduces a \emph{blind-spot} in the receptive field leads to an attenuation at very high frequencies. The number of iterations for LR is indicated, for each of the three images we choose the number of iterations with the best performance. Doted lines are displayed to help the eyes compare the spectra across different methods.}
  \label{fig:experiments}
\end{figure*}

\paragraph{Benchmark dataset.}
We tested the deconvolution performance of our model on a diverse set of $22$ two-dimensional monochrome images ranging in size between $512\times512$ and $2592\times1728$ pixels. The 22 images are normalised within $[0, 1]$. For each image we apply a Gaussian-like blur kernel\footnote{Corresponding to the optical point-spread-function of a $0.8NA$ $16\times$ microscope objective with $0.406\times0.406$ micron pixels.} followed by a Poisson-Gaussian noise model augmented by salt-and-pepper noise:
\begin{dmath}\label{eqn:noise_model}
  n(z) = s_p(z + \eta(z)N) 
\end{dmath}
Where $\eta(z) = \sqrt{\alpha z+\sigma^2}$, $\alpha$ is the Poisson term and $\sigma$ is the standard deviation of the Gaussian term, and $N$ is the independent normal Gaussian noise. Function $s_p$ applies 'salt-and-pepper' noise by replacing a proportion $p$ of pixels with a random value chosen uniformly within $[0, 1]$. In our experiments we choose a strong noise regime with: $\alpha = 0.001$, $\sigma = 0.1$, $p = 0.01$. Finally, the images are quantized with 10 bits of precision -- another source of image quality degradation.

\paragraph{Single image training.}
In true self-supervised fashion, we decided to train one model per image and not use any additional images for training. Adding more \emph{adequate} training instances or simply training on larger images would certainly help as the deep learning literature attests \cite{cho_how_2015}. However, here we are interested in the baseline performance in the purely self-supervised case. We do not generate training batches by tilling the images, but instead simply generate batches from a single image by sampling multiple random masks. Moreover, to further aid convergence, and guarantee sufficient stochasticity despite training on a single image, we use our own variant of the ADAM algorithm \cite{kingma2014adam} that adds epoch-decreasing normal gradient noise (see code repository for more details, link below). 

\paragraph{Results.}
\begin{figure*}[!ht]
  \centering
  \includegraphics[scale=1]{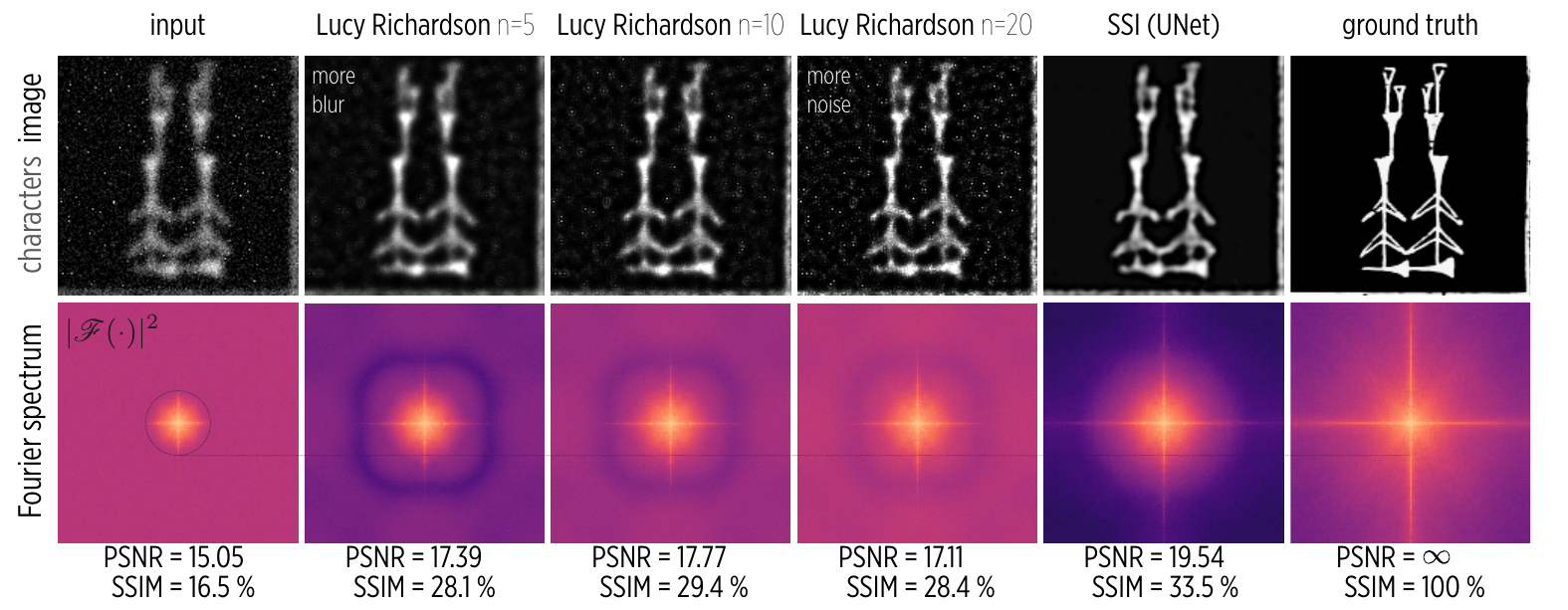}
  \caption{Lucy-Richardson deconvolution blur-noise tradeoff. As an iterative algorithm Lucy-Richardson deconvolution \cite{richardson1972bayesian} first reconstructs the image's low frequencies and then incrementally refines the reconstruction with higher frequency components. It follows that low-iteration reconstructions are less sensitive to noise whereas high-iteration reconstructions are sharper but also noisier -- hence a trade-off between sharpness and noise. In contrast, our self-supervised inversion approach is both insensitive to noise and sharpens the image.}
  \label{fig:lucyrich}
\end{figure*}

\paragraph{Comparison with classic approaches.}
We compare our Self-Supervised Inversion (SSI) approach with standard inversion algorithms such as Lucy-Richardson (LR) deconvolution \cite{richardson1972bayesian}, Conjugate Gradient optimization with TV prior \cite{chambolle2011first}, and Chambole-Pock primal-dual inversion also with a TV prior \cite{chambolle2011first}. In the case of LR deconvolution we evaluate image quality for three different number of iterations ($5$, $10$, and $20$) to explore the trade-off between noise amplification and sharpening (See Fig.~\ref{fig:lucyrich}).

\begin{table}
\caption{Average deconvolution performance per method for a benchmark set of $22$ images. We evaluate image fidelity between the ground truth and: blurry, blurry\&noisy, and restored images. The blurry\&noisy images are used as input for the different deconvolution methods. We compute the Peak Signal to Noise Ratio (PSNR) \cite{yuanji_image_2003}, Structural Similarity (SSIM) \cite{wang_multiscale_2003}, Mutual Information (MI) \cite{russakoff_image_2004}, and Spectral Mutual Information (SMI). For all metrics, higher is better. The metrics SSIM, MI, and SMI are always within $[0, 1]$ with $0$ being the worst value, and $1$ attained when the two images are identical. For all fidelity metrics image deconvolution by Self-Supervised Inversion performed best. }
\label{table:metrics}
\vskip 0.15in
\begin{center}
\begin{footnotesize}
\begin{filecontents*}{tables/scores_summary_best.csv}
method, PSNR, SSIM, MI, SMI
blurry, 23.1, 0.77, 0.17, 0.38
blurry\&noisy (input), 17.8, 0.29, 0.07, 0.18
Conjugate Gradient TV, 19.4, 0.41, 0.09, 0.21
Chambole Pock TV , 18.7, 0.40, 0.07, 0.23
Lucy Richardson $n=5$, 22.2, 0.59, 0.12, 0.25
Lucy Richardson $n=10$, 21.1, 0.52, 0.10, 0.25
Lucy Richardson $n=20$, 18.5, 0.38, 0.08, 0.19
SSI UNet \emph{no masking}, 17.7, 0.38, 0.07, 0.14
SSI UNet, \textbf{22.5}, \textbf{0.61}, \textbf{0.14}, \textbf{0.27}
\end{filecontents*}
\pgfplotstabletypeset[
    col sep=comma,
    string type,
    every head row/.style={%
        after row=\hline
    },
    columns/method/.style={column name= , column type=l},
    columns/PSNR/.style={column name=PSNR, column type=c},
    columns/SSIM/.style={column name=SSIM, column type=c},
    columns/MI/.style={column name=MI, column type=c},
    columns/SMI/.style={column name=SMI, column type=c},
    ]{tables/scores_summary_best.csv}
\end{footnotesize}
\end{center}
\vskip -0.1in
\end{table}

Table.~\ref{table:metrics} gives averages for four image comparison metrics: Peak Signal to Noise Ratio (PSNR) \cite{yuanji_image_2003}, Structural Similarity (SSIM)  \cite{wang_multiscale_2003}, Mutual Information (MI) \cite{russakoff_image_2004}, and Spectral Mutual Information (SMI). The SMI metric is novel and is designed to directly measure image fidelity in the frequency domain: we compute the mutual information of the the Discrete Cosine Transform (DCT~2) of both images. Different image comparison metrics have  different biases, hence we found important to evaluate several metrics to gain confidence on our results.

A selection of deconvolved image crops: \emph{target}, \emph{embryo}, \emph{characters} are shown in Fig.~\ref{fig:experiments}. We show the images as well as their Fourier spectra and compare the methods: Chambole-Pock, Lucy-Richardson, Self-Supervised Inversion, and a control: Self-Supervised Inversion without masking. Overall, we find that Self-Supervised Inversion achieves the best performance across all metrics evaluated with PSNR=$22.5$, SSIM=$0.61$, MI=$0.14$, and SMI=$0.27$. After SSI, the second best approach is Lucy-Richardson with $5$ iterations. However, visual inspection of the corresponding images and spectra shows that while these images have little noise they also lack sharpness (see Fig.~\ref{fig:lucyrich}). Different metrics will weight differently image differences due to noise, and image differences due to sharpness. Since sharpness only manifests itself along the edges of an image, whereas noise is present everywhere, it is expected that most metrics will favour noise reduction versus sharpness.

We were pleased to observe that SSI \emph{without masking} -- while producing worse images -- performed better than expected, or at least did not lead to excessive noise amplification. A possible explanation is that after applying the forward model $g$ the result cannot be noisy -- because  $g$ is a low-pass filter. The loss will be affected by the noise and disrupt training but to a lesser extent than in the Noise2Self \cite{batson2019noise2self} case where there is nothing to prevent the noise to propagate all the way through.

We also observed that images with repetitive and stereotypical patterns tend to have better self-supervised deconvolution quality. This is expected. For example, complex images where each image patch is distinct will not fare well with content aware methods \cite{weigert2018content, belthangady2019applications}, since we train on single images, the restoration quality will depend on how much redundancy can be found across the image. The less redundancy the more training data will be needed.

\begin{table}
\caption{Average inversion speed per method for a benchmark set of $22$ images. Note: Conjugate Gradient and Chambole Pock are implemented on CPU, whereas all other method are GPU accelerated.}
\label{table:timmings}
\vskip 0.15in
\begin{center}
\begin{footnotesize}
\begin{filecontents*}{tables/timming_summary_best.csv}
method, training time, inference time
Conjugate Gradient TV, 0.00, 95.74
Chambole Pock TV , 0.00, 306.60
Lucy Richardson $n=5$, 0.00, 0.23
Lucy Richardson $n=10$, 0.00, 0.10
Lucy Richardson $n=20$, 0.00, 0.17
SSI UNet \emph{no masking}, 222.67, 0.03
SSI UNet, 249.01, 0.03
\end{filecontents*}
\pgfplotstabletypeset[
    col sep=comma,
    string type,
    every head row/.style={%
        after row=\hline
    },
    columns/method/.style={column name=method, column type=l},
    columns/training time/.style={column name=training time (s), column type=c},
    columns/inference time/.style={column name=inference time (s), column type=c},
    ]{tables/timming_summary_best.csv}
\end{footnotesize}
\end{center}
\vskip -0.1in
\end{table}

Table \ref{table:timmings} lists the average training and inference times for different methods. Conjugate Gradient and Chambole Pock methods don't require training but are also very slow. As expected, CNN based methods have long training times but are capable of nearly instantaneous inference \footnote{$30$ ms, on a RTX TITAN GPU.}.

\section{Conclusion}

We have shown how to generalise the theory of $\J$-invariance from denoising to solving inverse problems in the presence of noise. Our theory is general: it applies to any reasonably posed inverse problem, does not require prior on the signal or noise, nor does it rely on clean training data. From an implementation standpoint, all that is needed is differentiable inverse and forward models. We have shown how noisy and blurry images can be restored individually -- without extra training data -- to image quality levels that surpass the classical inversion algorithms typically applied to deconvolution such as Lucy-Richardson deconvolution.

We are looking forward to extend this work to the 3D case and demonstrate noise-tolerant deconvolution of real microscopy data. The problem of blind inversion is the natural next step, but a much more difficult one. We are particularly interested in finding out if, again, there is a way to solve this problem without any clean training data or prior.

\section{Code and Methodological Details.}

Our Python implementation of Self-Supervised Inversion in PyTorch \cite{paszke_automatic_2017} with examples can be found here: \href{https://github.com/royerlab/ssi-code}{github.com/royerlab/ssi-code}. All details and parameters are provided with the code. The latest version of this document can be found \href{https://github.com/royerlab/ssi-code/blob/master/paper/Noise_Tolerant_Self_Supervised_Inversion.pdf}{here}.

\section*{Acknowledgements}

Thank you to the \emph{Chan Zuckerberg Biohub} for financial support. Loic A. Royer thanks his wonderfull wife Zana Vosough for letting him finish this work on a week-end.


\bibliography{main}

\begin{thebibliography}{30}
\providecommand{\natexlab}[1]{#1}
\providecommand{\url}[1]{\texttt{#1}}
\expandafter\ifx\csname urlstyle\endcsname\relax
  \providecommand{\doi}[1]{doi: #1}\else
  \providecommand{\doi}{doi: \begingroup \urlstyle{rm}\Url}\fi

\bibitem[Batson \& Royer(2019)Batson and Royer]{batson2019noise2self}
Batson, J. and Royer, L.
\newblock Noise2self: Blind denoising by self-supervision.
\newblock \emph{arXiv preprint arXiv:1901.11365}, 2019.

\bibitem[Belthangady \& Royer(2019)Belthangady and
  Royer]{belthangady2019applications}
Belthangady, C. and Royer, L.~A.
\newblock Applications, promises, and pitfalls of deep learning for
  fluorescence image reconstruction.
\newblock \emph{Nature methods}, pp.\  1--11, 2019.

\bibitem[Chambolle \& Pock(2011)Chambolle and Pock]{chambolle2011first}
Chambolle, A. and Pock, T.
\newblock A first-order primal-dual algorithm for convex problems with
  applications to imaging.
\newblock \emph{Journal of mathematical imaging and vision}, 40\penalty0
  (1):\penalty0 120--145, 2011.

\bibitem[Cho et~al.(2015)Cho, Lee, Shin, Choy, and Do]{cho_how_2015}
Cho, J., Lee, K., Shin, E., Choy, G., and Do, S.
\newblock How much data is needed to train a medical image deep learning system
  to achieve necessary high accuracy.
\newblock \emph{arXiv: Learning}, 2015.

\bibitem[Dong et~al.(2014)Dong, Loy, He, and Tang]{dong2014learning}
Dong, C., Loy, C.~C., He, K., and Tang, X.
\newblock Learning a deep convolutional network for image super-resolution.
\newblock In \emph{European conference on computer vision}, pp.\  184--199.
  Springer, 2014.

\bibitem[He et~al.(2019)He, Fan, Wu, Xie, and Girshick]{he2019momentum}
He, K., Fan, H., Wu, Y., Xie, S., and Girshick, R.
\newblock Momentum contrast for unsupervised visual representation learning.
\newblock \emph{arXiv preprint arXiv:1911.05722}, 2019.

\bibitem[Hendriksen et~al.(2020)Hendriksen, Pelt, and
  Batenburg]{hendriksen2020noise2inverse}
Hendriksen, A.~A., Pelt, D.~M., and Batenburg, K.~J.
\newblock Noise2inverse: Self-supervised deep convolutional denoising for
  linear inverse problems in imaging.
\newblock \emph{arXiv preprint arXiv:2001.11801}, 2020.

\bibitem[Kaneko \& Harada(2020)Kaneko and Harada]{kaneko2020blur}
Kaneko, T. and Harada, T.
\newblock Blur, noise, and compression robust generative adversarial networks.
\newblock \emph{arXiv preprint arXiv:2003.07849}, 2020.

\bibitem[Kingma \& Ba(2014)Kingma and Ba]{kingma2014adam}
Kingma, D.~P. and Ba, J.
\newblock Adam: A method for stochastic optimization.
\newblock \emph{arXiv preprint arXiv:1412.6980}, 2014.

\bibitem[Krishnan et~al.(2020)Krishnan, Belthangady, Nyby, Lange, Yang, and
  Royer]{krishnan2020optical}
Krishnan, A.~P., Belthangady, C., Nyby, C., Lange, M., Yang, B., and Royer,
  L.~A.
\newblock Optical aberration correction via phase diversity and deep learning.
\newblock \emph{bioRxiv}, 2020.

\bibitem[Krull et~al.(2019)Krull, Buchholz, and Jug]{krull2019noise2void}
Krull, A., Buchholz, T.-O., and Jug, F.
\newblock Noise2void-learning denoising from single noisy images.
\newblock In \emph{Proceedings of the IEEE Conference on Computer Vision and
  Pattern Recognition}, pp.\  2129--2137, 2019.

\bibitem[Laine et~al.(2019)Laine, Karras, Lehtinen, and Aila]{laine2019high}
Laine, S., Karras, T., Lehtinen, J., and Aila, T.
\newblock High-quality self-supervised deep image denoising.
\newblock In \emph{Advances in Neural Information Processing Systems}, pp.\
  6968--6978, 2019.

\bibitem[Lehtinen et~al.(2018)Lehtinen, Munkberg, Hasselgren, Laine, Karras,
  Aittala, and Aila]{lehtinen2018noise2noise}
Lehtinen, J., Munkberg, J., Hasselgren, J., Laine, S., Karras, T., Aittala, M.,
  and Aila, T.
\newblock Noise2noise: Learning image restoration without clean data.
\newblock \emph{arXiv preprint arXiv:1803.04189}, 2018.

\bibitem[McCann et~al.(2017)McCann, Jin, and Unser]{McCann2017Convolutional}
McCann, M.~T., Jin, K.~H., and Unser, M.
\newblock Convolutional neural networks for inverse problems in imaging: A
  review.
\newblock \emph{{IEEE} Signal Processing Magazine}, 34\penalty0 (6):\penalty0
  85--95, November 2017.
\newblock ISSN 1053-5888.
\newblock \doi{10.1109/MSP.2017.2739299}.

\bibitem[Misra \& van~der Maaten(2019)Misra and van~der Maaten]{misra2019self}
Misra, I. and van~der Maaten, L.
\newblock Self-supervised learning of pretext-invariant representations.
\newblock \emph{arXiv preprint arXiv:1912.01991}, 2019.

\bibitem[Moran et~al.(2019)Moran, Schmidt, Zhong, and
  Coady]{moran2019noisier2noise}
Moran, N., Schmidt, D., Zhong, Y., and Coady, P.
\newblock Noisier2noise: Learning to denoise from unpaired noisy data.
\newblock \emph{arXiv preprint arXiv:1910.11908}, 2019.

\bibitem[Mousavi \& Baraniuk(2017)Mousavi and Baraniuk]{mousavi2017learning}
Mousavi, A. and Baraniuk, R.~G.
\newblock Learning to invert: Signal recovery via deep convolutional networks.
\newblock In \emph{2017 IEEE international conference on acoustics, speech and
  signal processing (ICASSP)}, pp.\  2272--2276. IEEE, 2017.

\bibitem[Pajot et~al.(2018)Pajot, de~Bezenac, and
  Gallinari]{pajot2018unsupervised}
Pajot, A., de~Bezenac, E., and Gallinari, P.
\newblock Unsupervised adversarial image reconstruction.
\newblock 2018.

\bibitem[Paszke et~al.(2017)Paszke, Gross, Chintala, Chanan, Yang, DeVito, Lin,
  Desmaison, Antiga, and Lerer]{paszke_automatic_2017}
Paszke, A., Gross, S., Chintala, S., Chanan, G., Yang, E., DeVito, Z., Lin, Z.,
  Desmaison, A., Antiga, L., and Lerer, A.
\newblock Automatic differentiation in {PyTorch}.
\newblock In \emph{{NIPS}-{W}}, 2017.

\bibitem[Richardson(1972)]{richardson1972bayesian}
Richardson, W.~H.
\newblock Bayesian-based iterative method of image restoration.
\newblock \emph{JoSA}, 62\penalty0 (1):\penalty0 55--59, 1972.

\bibitem[Rick~Chang et~al.(2017)Rick~Chang, Li, Poczos, Vijaya~Kumar, and
  Sankaranarayanan]{rick2017one}
Rick~Chang, J., Li, C.-L., Poczos, B., Vijaya~Kumar, B., and Sankaranarayanan,
  A.~C.
\newblock One network to solve them all--solving linear inverse problems using
  deep projection models.
\newblock In \emph{Proceedings of the IEEE International Conference on Computer
  Vision}, pp.\  5888--5897, 2017.

\bibitem[Ronneberger et~al.(2015)Ronneberger, Fischer, and
  Brox]{ronneberger_u-net:_2015}
Ronneberger, O., Fischer, P., and Brox, T.
\newblock U-{Net}: Convolutional networks for biomedical image segmentation.
\newblock \emph{arXiv:1505.04597 [cs]}, May 2015.

\bibitem[Russakoff et~al.(2004)Russakoff, Tomasi, Rohlfing, and
  Maurer]{russakoff_image_2004}
Russakoff, D.~B., Tomasi, C., Rohlfing, T., and Maurer, C.~R.
\newblock Image similarity using mutual information of regions.
\newblock pp.\  596--607, 2004.

\bibitem[Ulyanov et~al.(2018)Ulyanov, Vedaldi, and Lempitsky]{ulyanov2018deep}
Ulyanov, D., Vedaldi, A., and Lempitsky, V.
\newblock Deep image prior.
\newblock In \emph{Proceedings of the IEEE Conference on Computer Vision and
  Pattern Recognition}, pp.\  9446--9454, 2018.

\bibitem[{Wang} et~al.(2003){Wang}, {Simoncelli}, and
  {Bovik}]{wang_multiscale_2003}
{Wang}, Z., {Simoncelli}, E.~P., and {Bovik}, A.~C.
\newblock Multiscale structural similarity for image quality assessment.
\newblock \emph{The Thrity-Seventh Asilomar Conference on Signals, Systems
  Computers, 2003}, 2:\penalty0 1398--1402 Vol.2, 2003.

\bibitem[{Wang Yuanji} et~al.(2003){Wang Yuanji}, {Li Jianhua}, {Lu Yi}, {Fu
  Yao}, and {Jiang Qinzhong}]{yuanji_image_2003}
{Wang Yuanji}, {Li Jianhua}, {Lu Yi}, {Fu Yao}, and {Jiang Qinzhong}.
\newblock Image quality evaluation based on image weighted separating block
  peak signal to noise ratio.
\newblock \emph{International Conference on Neural Networks and Signal
  Processing, 2003. Proceedings of the 2003}, 2:\penalty0 994--997 Vol.2, 2003.

\bibitem[Weigert et~al.(2018)Weigert, Schmidt, Boothe, M{\"u}ller, Dibrov,
  Jain, Wilhelm, Schmidt, Broaddus, Culley, et~al.]{weigert2018content}
Weigert, M., Schmidt, U., Boothe, T., M{\"u}ller, A., Dibrov, A., Jain, A.,
  Wilhelm, B., Schmidt, D., Broaddus, C., Culley, S., et~al.
\newblock Content-aware image restoration: pushing the limits of fluorescence
  microscopy.
\newblock \emph{Nature methods}, 15\penalty0 (12):\penalty0 1090--1097, 2018.

\bibitem[Xu et~al.(2014)Xu, Ren, Liu, and Jia]{xu2014deep}
Xu, L., Ren, J.~S., Liu, C., and Jia, J.
\newblock Deep convolutional neural network for image deconvolution.
\newblock In \emph{Advances in neural information processing systems}, pp.\
  1790--1798, 2014.

\bibitem[Zhang et~al.(2017)Zhang, Zuo, Chen, Meng, and
  Zhang]{zhang_beyond_2017}
Zhang, K., Zuo, W., Chen, Y., Meng, D., and Zhang, L.
\newblock Beyond a {G}aussian denoiser: Residual learning of deep {C}{N}{N} for
  image denoising.
\newblock \emph{IEEE Transactions on Image Processing}, 26\penalty0
  (7):\penalty0 3142--3155, July 2017.

\bibitem[Zhussip et~al.(2019)Zhussip, Soltanayev, and
  Chun]{zhussip2019training}
Zhussip, M., Soltanayev, S., and Chun, S.~Y.
\newblock Training deep learning based image denoisers from undersampled
  measurements without ground truth and without image prior.
\newblock In \emph{Proceedings of the IEEE Conference on Computer Vision and
  Pattern Recognition}, pp.\  10255--10264, 2019.

\end{thebibliography}
\bibliographystyle{icml2019}

\end{document}